\documentclass[a4paper,fleqn]{cas-dc}

\usepackage[numbers]{natbib}
\usepackage{setspace}
\usepackage{graphicx}
\usepackage{subcaption}
\definecolor{mygreen}{RGB}{34,139,34}

\def\tsc#1{\csdef{#1}{\textsc{\lowercase{#1}}\xspace}}
\tsc{WGM}
\tsc{QE}
\tsc{EP}
\tsc{PMS}
\tsc{BEC}
\tsc{DE}
\usepackage{algorithm}
\usepackage{algorithmic}

\begin{document}
\let\WriteBookmarks\relax
\def\floatpagepagefraction{1}
\def\textpagefraction{.001}
\shorttitle{Tackling Double Heterogeneity in Federated SHM}
\shortauthors{Y.H. Feng et~al.}

\title [mode = title]{Hierarchical Federated Learning with Dynamic Clustering and Adaptive Regularization for Robust Infrastructure Inspection}                      



\author[1]{Yuhu Feng}[orcid=0009-0006-4819-2066]
\credit{Conceptualization, Methodology, Software, Data curation, Writing - Original Draft}

\author[2]{Keisuke Maeda}[orcid=0000-0001-8039-3462]
\credit{Conceptualization, Supervision, Writing - Review \& Editing, Funding acquisition}

\author[2]{Takahiro Ogawa}[orcid=0000-0001-5332-8112]
\credit{Supervision, Writing - Review \& Editing, Project administration, Funding acquisition}

\author[2]{Miki Haseyama}[orcid=0000-0003-1496-1761]
\cormark[1]
\ead{mhaseyama@lmd.ist.hokudai.ac.jp} 
\credit{Supervision, Funding acquisition, Resources}

\affiliation[1]{organization={Graduate School of Information Science and Technology, Hokkaido University},
                addressline={North 14, West 9, Kita-ku}, 
                city={Sapporo},
                postcode={060-0814}, 
                state={Hokkaido},
                country={Japan}}

\affiliation[2]{organization={Faculty of Information Science and Technology, Hokkaido University},
                addressline={North 14, West 9, Kita-ku}, 
                city={Sapporo},
                postcode={060-0814}, 
                state={Hokkaido},
                country={Japan}}

\cortext[cor1]{Corresponding author}


\begin{abstract}
The deployment of data-driven computer vision models for structural health monitoring (SHM) is heavily constrained by the data silo dilemma due to stringent privacy and security regulations. While federated learning (FL) offers a privacy-preserving collaborative alternative, its application to nationwide infrastructure networks is severely hindered by the challenge of ``double heterogeneity'': macro-level physical divergence across disparate structural types and micro-level statistical imbalances within local datasets. To overcome this challenge, this paper proposes a novel hierarchical federated learning framework. The framework orchestrates a synergistic two-tier optimization strategy. At the macro-level, a dynamic gradient-based clustering mechanism autonomously aggregates distributed clients into specialized expert groups based on their structural degradation trajectories, circumventing the need for prior geographical metadata. Concurrently, at the micro-level, an intra-cluster Dynamic Region-Adaptive Proximal Regularization (DRAPR) module computes a real-time statistical Non-IID Intensity Score for each client. By adaptively modulating a proximal penalty based on local label skewness and gradient divergence, DRAPR effectively calibrates local updates, mitigates client drift, and prevents the catastrophic forgetting of minority damage classes. Comprehensive evaluations on a large-scale, real-world structural inspection dataset demonstrate that the hierarchical integration of macro-clustering and micro-regularization successfully neutralizes dual-level heterogeneity, yielding highly robust and specialized diagnostic models for complex infrastructure inspection.
\end{abstract}

\begin{graphicalabstract}
\includegraphics[clip,scale=0.47]{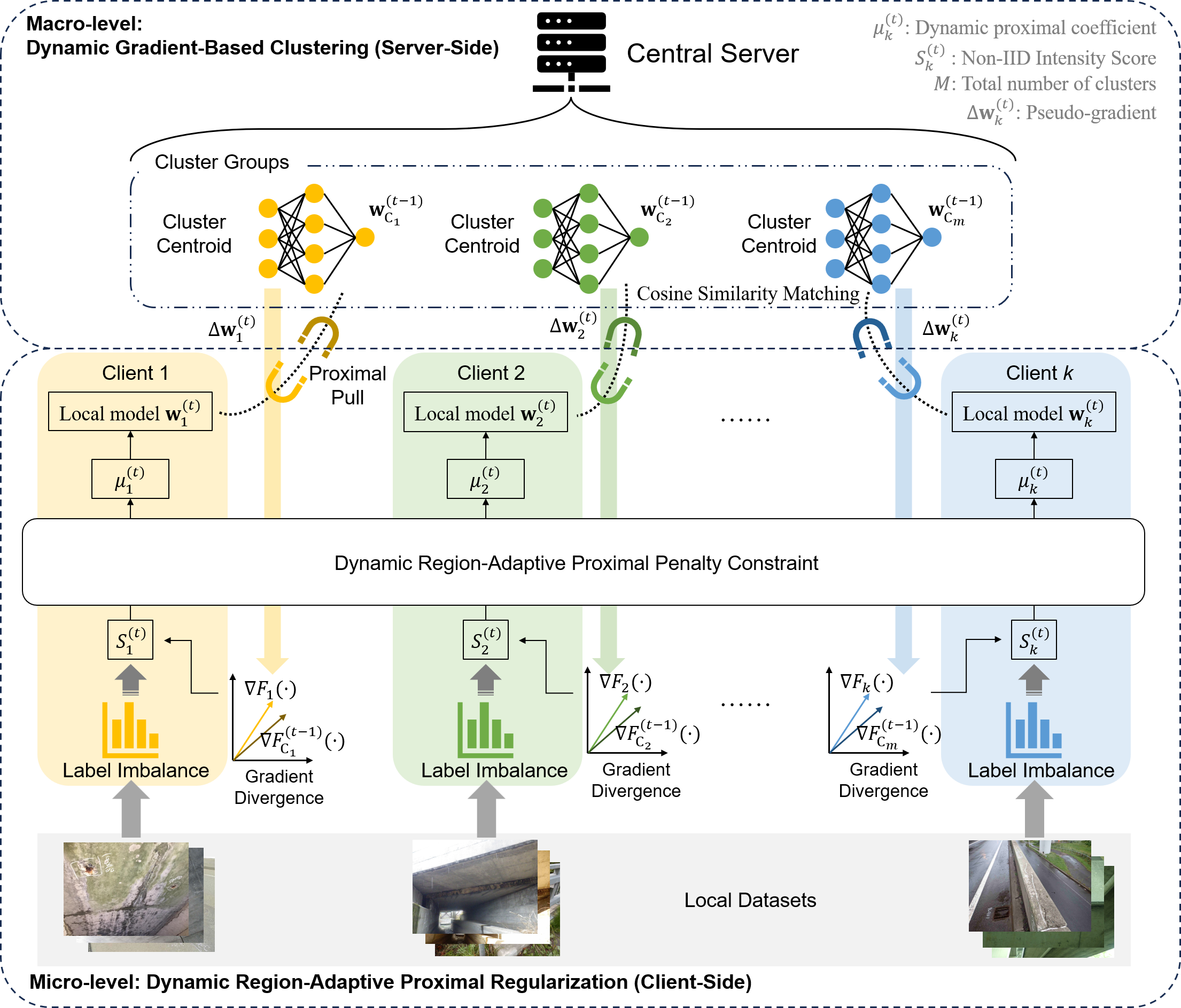}
\end{graphicalabstract}

\begin{highlights}
\item Clustered-DRAPR tackles double heterogeneity in federated infrastructure inspection.
\item Gradient clustering discovers macro-climatic damage patterns without metadata.
\item Dynamic adaptive regularization mitigates client drift from imbalanced labels.
\end{highlights}

\begin{keywords}
Structural Health Monitoring  \sep 
Federated Learning \sep 
Data Heterogeneity  \sep 
Multi-label Classification \sep 
Dynamic Clustering \sep 
Deep Learning
\end{keywords}

\maketitle


\section{Introduction}

The maintenance and management of civil infrastructure systems, particularly highway bridges and tunnels, represent critical global challenges due to the rapid aging of existing assets. 
Automated inspection systems based on computer vision and deep learning have emerged as promising solutions to address the labor shortages and subjectivity associated with manual visual inspections \cite{jia2023deep, lehman2022american, payawal2023image, Keisuke2019_6, Yuya2024_6}. 
State-of-the-art neural architectures have demonstrated remarkable success in detecting structural defects such as cracks, corrosion, and spalling \cite{el2025review, mohammadi2024advancements, Keisuke2024_11}. 
However, the deployment of these data-driven models in real-world engineering scenarios is severely hindered by the data silo problem. 
While local road management agencies and inspection organizations possess vast amounts of structural health monitoring (SHM) data \cite{bao2019state,sohn2003review, catbas2008structural,farrar2007introduction}, stringent privacy regulations and security concerns often prohibit the centralized aggregation of this sensitive information \cite{pan2025study, plevris2024ai}. 
Consequently, training a robust, generalized global model using fragmented data distributed across multiple institutions remains a formidable obstacle.

Federated Learning (FL) \cite{li2020review, zhang2021survey,kairouz2021advances} has been introduced as a privacy-preserving paradigm to circumvent these data barriers, allowing multiple clients to collaboratively train a model without sharing raw images \cite{mcmahan2017communication}. 
Despite its potential, the application of FL in civil infrastructure engineering \cite{anaissi2023personalised, anaissi2021intelligent} faces a unique and severe challenge: extreme and complex data heterogeneity (Non-IID data) at multiple levels \cite{kairouz2021advances, zhao2018federated}. 
To systematically address this, we conceptualize data heterogeneity into two distinct tiers. First, at the macro-level (inter-client), infrastructure data exhibits profound physical divergence. 
For instance, images from coastal steel bridges dominated by salt-induced corrosion differ fundamentally from those of inland concrete structures featuring freeze-thaw cracks \cite{ghosh2020efficient, sattler2020clustered}. 
Forcing all such diverse clients to converge to a single one-size-fits-all global model, as commonly practiced in standard FedAvg \cite{mcmahan2017communication} or FedProx \cite{zhao2018federated}, inevitably leads to suboptimal performance in specific domains.
Second, at the micro-level (intra-client), even among clients monitoring similar structures, the local training processes are highly unstable due to statistical imbalances in damage frequencies, varying lighting conditions, and sensor noise. 
Existing Clustered FL approaches attempt to address the macro-level issue by grouping clients \cite{mayakannan2023navigating, cowlishaw2025balancing, qin2023personalized}, but they typically rely on static assignments and revert to standard averaging within the clusters, entirely neglecting the micro-level training instability (client drift) that persists within each group \cite{chen2024classifier, jothimurugesan2023federated}.

To simultaneously tackle these dual-level heterogeneities, this paper proposes Clustered-DRAPR, a novel Hierarchical Dynamic Region-Adaptive Federated Learning framework tailored for infrastructure inspection networks. 
Diverging from conventional paradigms that rely on a single global model or static predefined groups, our framework orchestrates a synergistic two-tier optimization strategy. 
At the macro-level, we introduce a gradient-based dynamic clustering mechanism that automatically identifies and segregates clients with similar structural damage patterns into specialized expert groups, without requiring prior knowledge of their geographic or physical metadata. 
At the micro-level, to address the residual heterogeneity within each cluster, we propose a novel Dynamic Region-Adaptive Proximal Regularization (DRAPR) algorithm. 
As a core component of our framework, DRAPR continuously evaluates a statistical Non-IID intensity score for each client, which is derived from label imbalance and local gradient divergence, to dynamically modulate the regularization weight at every communication round.
This ensures that individual clients optimize effectively without diverging from their respective cluster centroids.

The main contributions of this paper are summarized as follows:
\begin{itemize}
    \item We propose Clustered-DRAPR, a systematic, hierarchical federated learning framework specifically designed to overcome the ``double heterogeneity" (inter-group structural differences and intra-group statistical variations) inherent in large-scale civil infrastructure monitoring.
    \item We introduce a dynamic clustering algorithm that adaptively groups distributed inspection clients based on the similarity of their evolving model updates, enabling the collaborative training of specialized models for distinct infrastructure types.
    \item We design the DRAPR algorithm as a novel intra-cluster optimization module. By dynamically adjusting the proximal regularization penalty according to real-time client heterogeneity, DRAPR effectively mitigates local client drift and stabilizes the training process.
    \item We conduct extensive experiments on a real-world national road inspection dataset. The results demonstrate that the proposed unified framework significantly accelerates convergence and outperforms state-of-the-art FL baselines in terms of classification accuracy, particularly under extreme Non-IID conditions.
\end{itemize}

The remainder of this paper is organized as follows. 
Section \ref{sec:relatedwork} provides a brief overview of related work. 
Section \ref{sec:method} presents a detailed description of the proposed method. 
The experimental results are presented in Section \ref{sec:experiment}, where we provide qualitative and quantitative evaluations of the proposed framework. 
Section \ref{sec:discussion} discusses the implications of our findings and the limitations associated with our study. 
Finally, Section \ref{sec:conclusion} concludes the paper.

\section{Related Works}
\label{sec:relatedwork}

\subsection{Deep Learning for SHM and Privacy Constraints}

In recent years, the integration of DL with computer vision has revolutionized the field of SHM, offering a robust alternative to labor-intensive manual inspections. Comprehensive reviews indicate that automated algorithms have significantly improved the efficiency of maintenance strategies for industrial machines and civil infrastructure \cite{mohammadi2024advancements, chun2022deep, maharjan2026large, yamane2023recording}. Specifically, state-of-the-art architectures, ranging from Convolutional Neural Networks (CNNs) to Vision Transformers (ViTs), have demonstrated exceptional performance in detecting structural defects. For instance, recent studies have successfully deployed DL models for the fine-grained classification of multi-level surface information in bridge inspections, achieving human-level accuracy \cite{zhang2024deep}. Similarly, in road maintenance, intelligent systems have been extensively developed to automatically identify and characterize pavement cracks under varying environmental conditions \cite{el2025review}.

However, the practical deployment of these centralized learning paradigms faces a formidable bottleneck: the data silo dilemma. As highlighted in recent research on AI-driven infrastructure safety, the effectiveness of these models relies heavily on the availability of large-scale, diverse training data \cite{plevris2024ai}. Yet, in the civil engineering sector, inspection data is typically fragmented across various stakeholders, such as regional transportation bureaus and private engineering firms. Furthermore, the emergence of digital twin technologies has heightened the complexity of data governance \cite{pan2025study}. Due to stringent data privacy regulations (e.g., GDPR) and concerns regarding the security of critical infrastructure assets, these entities are often reluctant or legally prohibited from sharing raw inspection imagery with a central server. Consequently, individual organizations are forced to train models on limited, isolated datasets, which inevitably leads to poor generalization performance. This conflict between the imperative of data-driven insights and the constraints of data privacy has catalyzed the adoption of decentralized learning frameworks.

\subsection{Federated Learning in Civil Infrastructure}

To address the privacy concerns inherent in cross-institutional data sharing, FL has been increasingly adopted in the civil engineering domain. Early implementations primarily utilized the standard FedAvg \cite{mcmahan2017communication} algorithm to aggregate local model updates, establishing the feasibility of decentralized training \cite{scarselli2025machine}. 
For instance, the authors in \cite{yang2025privacy} proposed a privacy-preserving framework specifically for miter gate monitoring, verifying that collaborative learning could be achieved without centralizing sensitive hydraulic infrastructure data. Similarly, other studies have begun to explore personalized FL schemes to better tailor damage detection models to specific structural contexts \cite{anaissi2021intelligent}. 
While these pioneering works have successfully validated the core utility of FL in SHM, they often operate under the assumption of relatively homogeneous data distributions or rely on basic aggregation strategies \cite{moshawrab2023reviewing, pillutla2022robust}. 
Consequently, they may struggle to maintain performance when applied to large-scale networks where inspection images exhibit severe statistical heterogeneity due to varying camera sensors and environmental backgrounds.

Recognizing the limitations of a single one-size-fits-all global model, recent research has started to investigate multi-model approaches. Notably, the study in \cite{cheema2025clustered} introduced a clustered FL framework for population-based SHM, which partitions the global population of structures into sub-groups to improve prediction accuracy. 
This approach aligns with the intuition that civil infrastructure naturally forms distinct categories (e.g., suspension bridges vs. truss bridges). However, existing clustered FL methods in this domain typically rely on static assignment mechanisms or hard clustering based on pre-defined physics-based features. 
They often lack the capability to dynamically adapt to the evolving gradient directions of deep neural networks during the training process. 
Furthermore, once clusters are formed, these methods usually revert to standard averaging within the group, ignoring the residual Non-IID issues among clients in the same cluster \cite{lu2024federated, briggs2020federated}. This highlights the need for a hierarchical framework that can simultaneously manage inter-group structural differences via dynamic clustering and intra-group statistical variations via adaptive regularization.

\subsection{Handling Non-IID Data: Regularization and Clustering}

To mitigate the performance degradation caused by statistical heterogeneity, recent advancements in the broader machine learning community have primarily focused on two strategies: regularization-based optimization and multi-center clustering.

In the realm of regularization, the seminal work FedProx \cite{li2020federated} introduced a proximal term to the local objective function, constraining local updates to remain close to the global model. By limiting the impact of variable local work, FedProx effectively tackles the statistical heterogeneity that often destabilizes standard FedAvg. However, FedProx employs a constant regularization coefficient across all clients and communication rounds. This static approach is suboptimal for infrastructure monitoring networks, where the degree of heterogeneity is dynamic and client-specific (e.g., varying lighting conditions or seasonal structural changes). To address the issue of client drift more explicitly, SCAFFOLD \cite{karimireddy2020scaffold} proposed a stochastic controlled averaging method. By utilizing control variates to estimate the update direction of the server and clients, SCAFFOLD corrects the drift in local updates, significantly accelerating convergence. Despite its efficacy, SCAFFOLD still fundamentally assumes that a single global model can generalize to all clients. In civil engineering scenarios exhibiting extreme heterogeneity (e.g., distinct damage patterns between suspension and masonry bridges), forcing convergence to a single centroid---even with drift correction---may compromise the model's ability to capture specialized structural features.

Recognizing that a single model may be insufficient, clustered FL allows clients to be grouped into multiple clusters, each maintaining its own specialized model. A representative approach, IFCA (Iterative Federated Clustering Algorithm) \cite{ghosh2020efficient}, alternately estimates cluster identities by minimizing the local loss and subsequently updates the cluster-specific models. While IFCA successfully captures the multi-modal nature of Non-IID data, it typically employs a hard assignment mechanism and reverts to standard averaging within each cluster. It lacks a fine-grained mechanism to handle the residual heterogeneity within a cluster, which is critical when clients in the same structural group still possess diverse sensor characteristics or data imbalances. This limitation underscores the necessity for the proposed Clustered-DRAPR framework, which integrates the macro-level specialization of clustering with the micro-level robustness of dynamic adaptive regularization.

\section{Methodology}
\label{sec:method}

\begin{figure*}[htbp]
    \centering
    \includegraphics[width=0.95\textwidth]{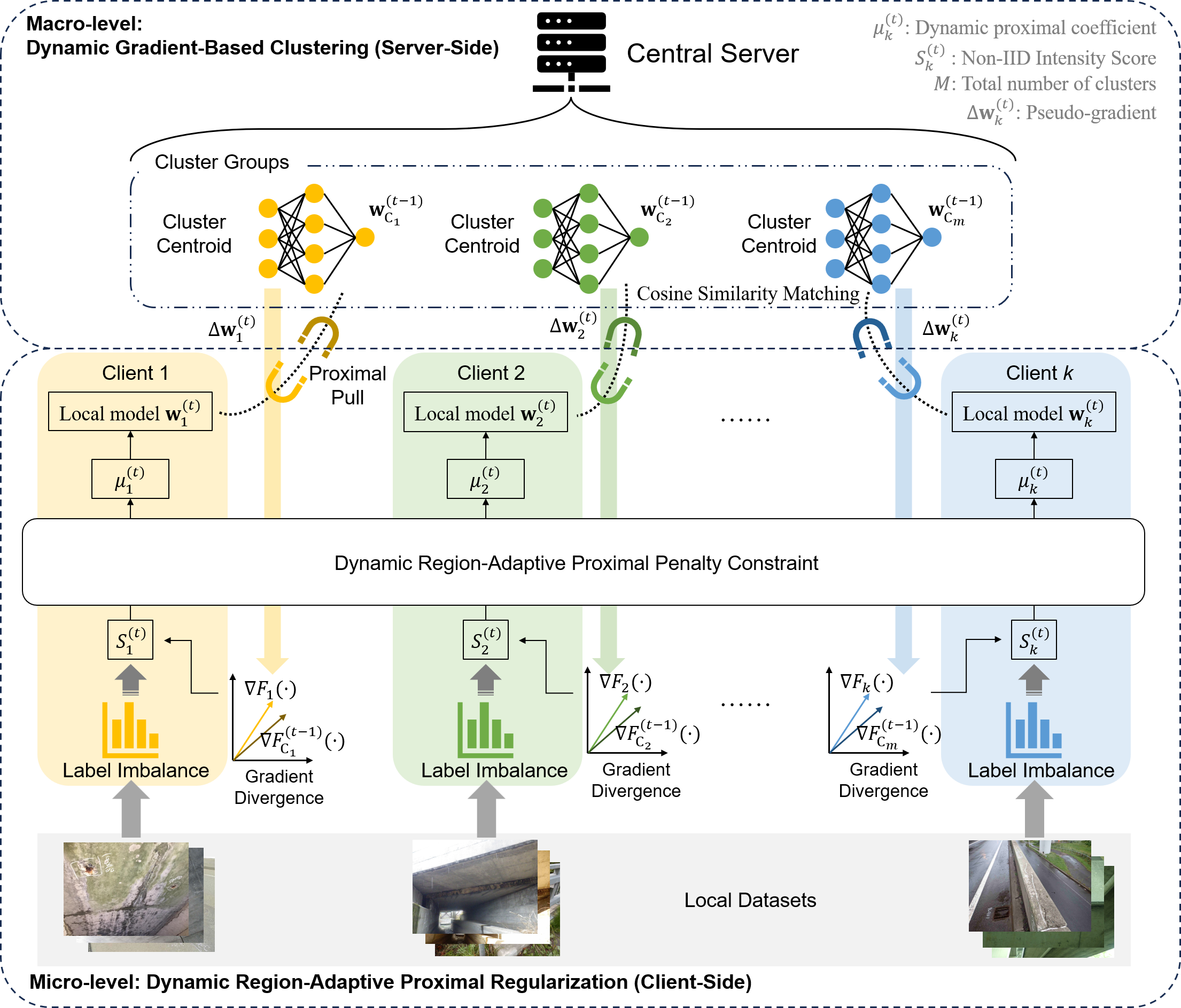} 
    \caption{The overall architecture of the proposed Clustered-DRAPR framework. This framework mitigates double heterogeneity through a synergistic two-tier strategy: (top) macro-level dynamic gradient-based clustering at the central server to form specialized expert models without raw data sharing, and (bottom) micro-level DRAPR at the client side to dynamically penalize local drift based on real-time data distribution skewness.}
    \label{fig:overview}
\end{figure*}

Before delving into the mathematical formulations, Figure \ref{fig:overview} illustrates the overall architecture of the proposed Clustered-DRAPR framework. This framework operates through a synergistic, dual-level optimization strategy to address the double heterogeneity inherent in decentralized infrastructure data. At the macro-level (server-side), a dynamic gradient-based clustering mechanism aggregates clients with similar structural degradation patterns into distinct expert clusters via cosine similarity matching. Simultaneously, at the micro-level (client-side), the DRAPR module computes a statistical Non-IID Intensity Score based on local label imbalance and gradient divergence. This score continuously modulates an adaptive proximal pull, preventing local models from drifting away from their respective cluster centroids during training.

\subsection{Problem Formulation and Preliminaries}
\label{subsec:3.1}
In the context of SHM, we consider a decentralized inspection network consisting of a set of clients $\mathcal{K} = \{1, 2, \dots, K\}$, where each client represents a regional road management agency or a specific infrastructure monitoring node. Each client $k \in \mathcal{K}$ holds a private, local dataset $\mathcal{D}_k = \{(x_i, y_i)\}_{i=1}^{N_k}$ containing $N_k = |\mathcal{D}_k|$ samples. Here, $x_i$ denotes an inspection image (e.g., a photograph of a bridge surface or tunnel lining), and $y_i \in \mathcal{Y}$ represents the corresponding structural condition or damage label (e.g., intact, crack, or corrosion).

The standard objective of FL is to collaboratively train a global machine learning model, parameterized by $\mathbf{w}$, without exposing the raw localized data $\mathcal{D}_k$. This is typically formulated as minimizing the global empirical risk function $L(\mathbf{w})$
\begin{equation}
    \min_{\mathbf{w}} L(\mathbf{w}) = \sum_{k=1}^{K} p_k F_k(\mathbf{w}),
    \label{eq:global_loss}
\end{equation}
where $p_k = \frac{N_k}{\sum_{j=1}^K N_j}$ determines the relative impact of client $k$ based on its data volume, such that $\sum_{k=1}^K p_k = 1$. The term $F_k(\mathbf{w})$ is the local empirical loss function for client $k$, defined as:
\begin{equation}
    F_k(\mathbf{w}) = \frac{1}{N_k} \sum_{(x_i, y_i) \in \mathcal{D}_k} \ell(\mathbf{w}; x_i, y_i),
\end{equation}
where $\ell(\cdot)$ is a task-specific loss function, such as the cross-entropy loss for damage classification tasks.

In traditional FL algorithms like FedAvg, clients perform multiple epochs of local stochastic gradient descent (SGD) to minimize $F_k(\mathbf{w})$, and a central server periodically aggregates these local updates to compute a single global model $\mathbf{w}^G$. This paradigm implicitly assumes that the data distribution across all clients is Independent and Identically Distributed (IID), meaning $P_i(x, y) = P_j(x, y)$ for any two clients $i$ and $j$. 

However, in real-world civil engineering applications, the IID assumption is strictly violated, leading to severe performance degradation. We formulate this challenge as a \textbf{Double Heterogeneity} problem, characterized by two levels of distributional divergence:
\begin{itemize}
    \item \textbf{Macro-level Feature Shift (Physical Divergence):} The marginal feature distributions vary significantly across clients, i.e., $P_i(x) \neq P_j(x)$. This is driven by varying structural topologies, construction materials, and environmental backgrounds. For example, a client monitoring coastal steel bridges will produce images vastly different in texture and lighting compared to a client monitoring inland concrete tunnels. A single global model $\mathbf{w}^G$ lacks the capacity to simultaneously represent these distinct physical domains optimally.
    \item \textbf{Micro-level Label Shift (Statistical Imbalance):} Even among clients monitoring the exact same type of infrastructure, the marginal label distributions diverge, i.e., $P_i(y) \neq P_j(y)$. Damage occurrences are statistically imbalanced; certain regions may experience a high frequency of corrosion due to climate, while others may rarely encounter it. During local training, this statistical imbalance causes the local objective $F_k(\mathbf{w})$ to deviate sharply from the global objective $L(\mathbf{w})$, a phenomenon known as \textit{client drift}. 
\end{itemize}

To overcome the inherent limitations of Equation (\ref{eq:global_loss}) in the presence of extreme double heterogeneity, we propose abandoning the single global model constraint. Instead, we formulate a hierarchical optimization problem that seeks to find a set of specialized cluster-centric models $\mathcal{W}_C = \{\mathbf{w}_{C_1}, \mathbf{w}_{C_2}, \dots, \mathbf{w}_{C_M}\}$, while dynamically constraining local client drift within each cluster.

\subsection{Inter-Cluster Optimization: Dynamic Gradient-Based Clustering}
\label{subsec:3.2}
To address the macro-level physical divergence described in Section \ref{subsec:3.1}, our framework abandons the single global model paradigm in favor of maintaining a set of $M$ specialized cluster centroids, denoted as $\mathcal{W}_C = \{\mathbf{w}_{C_1}, \mathbf{w}_{C_2}, \dots, \mathbf{w}_{C_M}\}$. The primary challenge in forming these clusters within a federated environment is the strict prohibition against accessing raw client data $\mathcal{D}_k$. To circumvent this privacy constraint, we leverage the clients' model updates as a secure proxy for their underlying data distributions. 

In deep neural networks for computer vision, it is well-established that the gradients computed during backpropagation encode rich semantic information about the training data. Clients monitoring infrastructure with similar damage patterns (e.g., two distinct coastal regions predominantly experiencing steel corrosion) will yield model updates that point in similar directions within the high-dimensional parameter space. Let $\Delta \mathbf{w}_k^{(t)}$ denote the pseudo-gradient, or the model update, of client $k$ at communication round $t$, defined as:
\begin{equation}
    \Delta \mathbf{w}_k^{(t)} = \mathbf{w}_k^{(t)} - \mathbf{w}_{C_{m(k)}}^{(t-1)},
\end{equation}
where $\mathbf{w}_k^{(t)}$ is the locally optimized model of client $k$, and $\mathbf{w}_{C_{m(k)}}^{(t-1)}$ is the centroid model of the cluster to which client $k$ was assigned in the previous round. 

To quantify the structural similarity between the data distributions of different clients, we utilize the cosine similarity metric. Unlike Euclidean distance, which can be overly sensitive to the magnitude of the updates (often skewed by local data volumes $N_k$), cosine similarity effectively captures the directional alignment of the high-dimensional weight vectors. The similarity between the updates of client $k$ and the reference update of cluster $m$ is computed at the server side as:
\begin{equation}
    \text{sim}(\Delta \mathbf{w}_k^{(t)}, \Delta \mathbf{w}_{C_m}^{(t)}) = \frac{\langle \Delta \mathbf{w}_k^{(t)}, \Delta \mathbf{w}_{C_m}^{(t)} \rangle}{\|\Delta \mathbf{w}_k^{(t)}\|_2 \|\Delta \mathbf{w}_{C_m}^{(t)}\|_2},
\end{equation}
where $\langle \cdot, \cdot \rangle$ denotes the inner product, and $\Delta \mathbf{w}_{C_m}^{(t)}$ represents the moving average of the gradient direction for cluster $m$.

Based on this similarity metric, the central server performs dynamic clustering at the end of each communication round. The assignment of client $k$ to a specific cluster $c_k^{(t)} \in \{1, \dots, M\}$ is determined by identifying the cluster centroid that exhibits the highest directional similarity with the client's local update:
\begin{equation}
    c_k^{(t)} = \arg\max_{m \in \{1, \dots, M\}} \text{sim}(\Delta \mathbf{w}_k^{(t)}, \Delta \mathbf{w}_{C_m}^{(t)}),
\end{equation}
This dynamic assignment mechanism ensures that the clustering is not static but actively adapts to the evolving representations of the DL model. A client may be reassigned to a different expert cluster if its local data distribution shifts (e.g., due to the introduction of new seasonal inspection data).

Once all clients $\mathcal{K}$ are partitioned into mutually exclusive cluster sets $\mathcal{K}_1^{(t)}, \mathcal{K}_2^{(t)}, \dots, \mathcal{K}_M^{(t)}$, the server aggregates the local models within each cluster independently to form the new specialized centroid models for the subsequent round:
\begin{equation}
    \mathbf{w}_{C_m}^{(t)} = \sum_{k \in \mathcal{K}_m^{(t)}} \frac{N_k}{\sum_{j \in \mathcal{K}_m^{(t)}} N_j} \mathbf{w}_k^{(t)}, \quad \forall m \in \{1, \dots, M\},
\end{equation}
Through this inter-cluster optimization process, our framework successfully isolates macro-level heterogeneity, allowing the system to cultivate a diverse set of expert models specifically tailored to the varied physical characteristics of the national infrastructure network.

\subsection{Intra-Cluster Optimization: Dynamic Region-Adaptive Proximal Regularization}
\label{subsec:3.3}

While the macro-level dynamic clustering detailed in Section \ref{subsec:3.2} successfully mitigates physical divergence by grouping structurally similar clients, severe statistical heterogeneity often persists \textit{within} each cluster. Even if a set of clients monitors the same type of infrastructure (e.g., concrete bridges), their local datasets may still exhibit extreme variations in damage frequencies and data volumes due to distinct local climates or inspection schedules. Such intra-cluster Non-IID data can induce client drift, where a local model rapidly overfits to its imbalanced data, deviating significantly from the cluster's consensus and destabilizing the aggregated centroid model. 

To counteract this intra-cluster heterogeneity, we introduce the novel DRAPR module. Inspired by the fixed proximal term in FedProx, which restricts local updates to remain close to the global model, DRAPR goes a step further by introducing an adaptive mechanism tailored to the real-time heterogeneity of each client within its specific cluster. The modified local objective function for client $k \in \mathcal{K}_m^{(t)}$ is formulated as:
\begin{equation}
    \mathcal{L}_k^{\rm adapt}(\mathbf{w}_k) = F_k(\mathbf{w}_k) + \frac{\mu_k^{(t)}}{2} \|\mathbf{w}_k - \mathbf{w}_{C_{m(k)}}^{(t-1)}\|^2,
    \label{eq:drapr_loss}
\end{equation}
where $F_k(\mathbf{w}_k)$ is the empirical loss on the local dataset $\mathcal{D}_k$, $\mathbf{w}_{C_{m(k)}}^{(t-1)}$ is the received centroid model of cluster $m(k)$ to which client $k$ was assigned in the previous round, and $\mu_k^{(t)}$ is the dynamic proximal regularization coefficient.

The core innovation of DRAPR lies in the dynamic computation of $\mu_k^{(t)}$. Rather than applying a uniform constraint across all clients, DRAPR quantifies a Non-IID Intensity Score $S_k^{(t)}$ for each client. This score is mathematically derived from two primary sources of local divergence: label imbalance and gradient variation.

First, we quantify the structural label imbalance inherent in the client's local dataset. Let $\text{imb}_k$ denote the label imbalance ratio, defined as:
\begin{equation}
    \text{imb}_k = \frac{\sigma_k}{\nu_k + \epsilon},
    \label{eq:imbalance}
\end{equation}
where $\sigma_k$ and $\nu_k$ represent the standard deviation and the mean of the damage label distribution across the local dataset $\mathcal{D}_k$, respectively, and $\epsilon$ is a small constant to ensure numerical stability. A higher $\text{imb}_k$ indicates a severely skewed dataset (e.g., predominantly intact images with extremely rare crack samples).

Second, we measure the local gradient divergence, which reflects how far the client's optimization trajectory deviates from the cluster's collective direction. By combining these two metrics, the comprehensive Non-IID Intensity Score $S_k^{(t)}$ is computed as:
\begin{equation}
    S_k^{(t)} = \alpha \|\nabla F_k(\mathbf{w}_k) - \nabla F_{C_{m(k)}}(\mathbf{w}_{C_{m(k)}}^{(t-1)})\|^2 + \beta \cdot \text{imb}_k,
    \label{eq:intensity_score}
\end{equation}
where $\alpha$ and $\beta$ are weighting hyperparameters balancing the contribution of gradient divergence and label imbalance.

Finally, utilizing the computed score, the dynamic proximal coefficient $\mu_k^{(t)}$ for the current communication round is adaptively scaled:
\begin{equation}
    \mu_k^{(t)} = \mu_0 \cdot \frac{S_k^{(t)}}{\overline{S}_{C_m}^{(t-1)} + \epsilon},
    \label{eq:dynamic_mu}
\end{equation}
where $\mu_0$ acts as a base regularization hyperparameter, and $\overline{S}_{C_m}^{(t-1)}$ is the average Non-IID Intensity Score of all clients assigned to cluster $m$ from the previous communication round.
To ensure the stability of the optimization process, the calculation of the Non-IID Intensity Score $S_k^{(t)}$ is performed at a precise temporal junction. Specifically, immediately after the central server broadcasts the cluster prototype $\mathbf{w}_{C_{m(k)}}^{(t-1)}$ to client $k$, and prior to the commencement of the $E$ local optimization epochs, the client triggers the computation of $S_k^{(t)}$ and the subsequent update of $\mu_k^{(t)}$. This proactive calibration ensures that the regularization intensity is synchronized with the initial state of the local model at the start of each communication round, effectively anchoring the entire local optimization trajectory without introducing intra-round fluctuations in the penalty strength.

From an infrastructure engineering perspective, this mechanism functions as an intelligent anchor. If a regional node possesses a highly skewed local dataset—for instance, observing only severe corrosion while completely lacking samples of minor cracks—its calculated $S_k^{(t)}$ will be notably high. Consequently, Equation (\ref{eq:dynamic_mu}) will assign a larger regularization weight $\mu_k^{(t)}$, exerting a stronger mathematical pull that prevents the local model from catastrophically forgetting the shared knowledge of minor cracks learned by the cluster centroid. Conversely, clients with well-balanced, representative datasets are granted more flexibility (a lower $\mu_k^{(t)}$) to fine-tune the model, thereby jointly accelerating convergence and enhancing the generalizability of the cluster expert models.

\subsection{The Unified Framework: Clustered-DRAPR}

Integrating the macro-level dynamic clustering mechanism with the micro-level DRAPR module, we present the complete Hierarchical Dynamic Region-Adaptive Federated Learning framework, termed Clustered-DRAPR. The workflow seamlessly orchestrates the collaboration between the central server and the distributed infrastructure nodes without requiring any exchange of raw inspection images.

A typical communication round $t$ within the Clustered-DRAPR framework executes the following four sequential phases:
\begin{enumerate}
    \item \textbf{Broadcasting (Server-Side):} The central server transmits the latest cluster centroid models $\mathcal{W}_C^{(t-1)}$ to their respectively assigned clients based on the clustering results from the previous round.
    \item \textbf{Local DRAPR Optimization (Client-Side):} Upon receiving its designated centroid model $\mathbf{w}_{C_{m(k)}}^{(t-1)}$, each client $k$ evaluates its real-time Non-IID Intensity Score $S_k^{(t)}$ locally using Equation (\ref{eq:intensity_score}). Based on this score, the client calculates the dynamic proximal coefficient $\mu_k^{(t)}$. Subsequently, the client performs $E$ epochs of local training to minimize the adaptive objective function $\mathcal{L}_k^{adapt}$ (Equation (\ref{eq:drapr_loss})). The resulting pseudo-gradient $\Delta \mathbf{w}_k^{(t)}$ and the intensity score $S_k^{(t)}$ are then securely transmitted to the server.
    \item \textbf{Dynamic Clustering (Server-Side):} The server collects the updates from all participating clients. Using the cosine similarity metric on the pseudo-gradients $\Delta \mathbf{w}_k^{(t)}$, the server re-evaluates the structural affinities of the clients and dynamically partitions them into $M$ distinct expert clusters for the current round.
    \item \textbf{Intra-Cluster Aggregation (Server-Side):} Within each newly formed cluster $m$, the server performs a weighted aggregation of the local models to update the centroid model $\mathbf{w}_{C_m}^{(t)}$.
\end{enumerate}

The complete step-by-step execution logic of the proposed framework is formalized in Algorithm \ref{alg:clustered_drapr}.
It is worth noting that the proposed Clustered-DRAPR framework is highly communication-efficient and computationally lightweight. The dynamic clustering strictly utilizes the pseudo-gradients $\Delta \mathbf{w}_k^{(t)}$ already required for standard federated aggregation, while the Non-IID Intensity Score $S_k^{(t)}$ is a negligible scalar value. Consequently, the framework introduces minimal communication overhead compared to the conventional FedAvg baseline, making it practically viable for deployment on edge devices located at remote infrastructure sites with limited bandwidth.

\begin{algorithm}[tbp]
\caption{The Proposed Clustered-DRAPR Framework}
\label{alg:clustered_drapr}
\begin{algorithmic}[1]
\REQUIRE Set of clients $\mathcal{K}$, total communication rounds $T$, local epochs $E$, learning rate $\eta$, base regularization weight $\mu_0$, number of clusters $M$.
\ENSURE A set of optimized expert cluster models $\mathcal{W}_C^{(T)} = \{\mathbf{w}_{C_1}^{(T)}, \dots, \mathbf{w}_{C_M}^{(T)}\}$.
\STATE \textbf{Server Initialization:} Initialize $M$ cluster centroid models $\mathbf{w}_{C_1}^{(0)}, \dots, \mathbf{w}_{C_M}^{(0)}$. Randomly assign clients to initial clusters.
\FOR{each round $t = 1, 2, \dots, T$}
    \STATE \textbf{Server:} Broadcast $\mathbf{w}_{C_{m(k)}}^{(t-1)}$ to each assigned client $k$.
    \FOR{each client $k \in \mathcal{K}$ \textbf{in parallel}}
        \STATE Compute Non-IID Intensity Score $S_k^{(t)}$ based on local label imbalance and gradient divergence.
        \STATE Compute dynamic coefficient $\mu_k^{(t)} = \mu_0 \cdot \frac{S_k^{(t)}}{\overline{S}_{C_m}^{(t-1)} + \epsilon}$.
        \STATE Initialize local model: $\mathbf{w}_k \leftarrow \mathbf{w}_{C_{m(k)}}^{(t-1)}$.
        \FOR{local epoch $e = 1$ to $E$}
            \STATE Compute local adaptive loss $\mathcal{L}_k^{adapt}$ using $\mu_k^{(t)}$.
            \STATE Update model: $\mathbf{w}_k \leftarrow \mathbf{w}_k - \eta \nabla \mathcal{L}_k^{adapt}(\mathbf{w}_k)$.
        \ENDFOR
        \STATE Calculate local pseudo-gradient: $\Delta \mathbf{w}_k^{(t)} = \mathbf{w}_k - \mathbf{w}_{C_{m(k)}}^{(t-1)}$.
        \STATE Transmit $\Delta \mathbf{w}_k^{(t)}$ and $S_k^{(t)}$ to the server.
    \ENDFOR
    \STATE \textbf{Server:} 
    \STATE Compute moving average direction $\Delta \mathbf{w}_{C_m}^{(t)}$ for each cluster.
    \FOR{each client $k \in \mathcal{K}$}
        \STATE Assign client $k$ to cluster $c_k^{(t)}$:
        \STATE \qquad $c_k^{(t)} = \arg\max_m \text{sim}(\Delta \mathbf{w}_k^{(t)}, \Delta \mathbf{w}_{C_m}^{(t)})$.
    \ENDFOR
    \FOR{each cluster $m = 1$ to $M$}
        \STATE Aggregate local models for cluster $m$:
        \STATE \qquad $\mathbf{w}_{C_m}^{(t)} = \sum_{k \in \mathcal{K}_m^{(t)}} \frac{N_k}{\sum_{j \in \mathcal{K}_m^{(t)}} N_j} \mathbf{w}_k^{(t)}$.
    \ENDFOR
\ENDFOR
\end{algorithmic}
\end{algorithm}

\section{Experiments}
\label{sec:experiment}

To comprehensively evaluate the effectiveness of the proposed Clustered-DRAPR framework, we conduct extensive experiments on a real-world, large-scale civil infrastructure dataset. This section details the dataset configuration, evaluation metrics, baseline methods, and implementation specifics.

\subsection{Experimental Setup and Datasets}

\noindent\textbf{Dataset and Client Partitioning:} 
The experiments utilize bridge inspection images and corresponding diagnostic records spanning from 2019 to 2021, sourced from the Japanese National Road Facility Inspection Database (xROAD) \cite{xroad2025}. 
To simulate a realistic FL scenario characterized by physical and statistical heterogeneity, we geographically partition the dataset into nine distinct clients corresponding to major administrative regions in Japan: Hokkaido, Tohoku, Kanto, Hokuriku, Chubu, Kinki, Chugoku, Shikoku, and Kyushu. 
In strict adherence to the privacy-preserving premise of FL, data is isolated within each region, and no raw images are shared among clients or with the central server. 
A total of 77,890 damage images were extracted for this study. The task is formulated as a multi-label image classification problem, where each image is annotated with one or more damage types (e.g., crack, corrosion, delamination). 
Within each of the nine regional clients, the local dataset is randomly split into a training set (90\%) and an evaluation set (10\%).

\noindent\textbf{Heterogeneity Settings:}
To rigorously assess the algorithm's robustness under varying degrees of task complexity and Non-IID intensity, we design two distinct experimental scenarios based on the number of target damage classes ($N_c$):
\begin{itemize}
    \item \textbf{5-Class Setting ($N_c=5$):} Focuses on the five most prevalent and critical structural defects (Label IDs 01-05: Corrosion, Crack, Exposed Rebar, Delamination, and Deformation). This represents a moderate level of data heterogeneity.
    \item \textbf{20-Class Setting ($N_c=20$):} Expands the diagnostic scope to include all 20 standardized damage types, incorporating 15 additional fine-grained and rare defects (e.g., Bearing Dysfunction, Debris Accumulation). This simulates a highly complex and severe Non-IID environment, as these rare damages exhibit extreme regional imbalance. A comprehensive list detailing all 20 damage categories is provided in Appendix \ref{appendix:labels}.
\end{itemize}

\noindent\textbf{Evaluation Metrics:}
The classification performance is evaluated using four standard metrics: Accuracy (Subset Accuracy for multi-label tasks), Precision, Recall, and the F1-score. Given the inherent class imbalance typical of structural damage datasets, all metrics are computed using a \textit{weighted average} approach. This ensures that the evaluation is not disproportionately skewed by the majority classes, providing a fair assessment of the model's predictive capability across all damage types.

\noindent\textbf{Baselines and Implementation Details:}
To demonstrate the superiority of the proposed Clustered-DRAPR, we benchmark it against the following state-of-the-art methods:
\begin{itemize}
    \item \textbf{Local Training (Local):} Each client trains an independent model using only its private dataset for 80 epochs, with absolutely no parameter exchange. This serves as the lower-bound baseline reflecting the data silo problem.
    \item \textbf{FedAvg \cite{mcmahan2017communication}:} The standard FL baseline utilizing unconstrained local updates and a single global model.
    \item \textbf{FedProx \cite{zhao2018federated}:} A regularization-based FL method applying a fixed proximal term to mitigate client drift.
    \item \textbf{SCAFFOLD \cite{karimireddy2020scaffold}:} An optimization-based FL method that utilizes control variates to correct gradient drift.
    \item \textbf{MOON \cite{li2021model}:} A contrastive learning-based FL method designed to align image features across clients.
    \item \textbf{IFCA \cite{ghosh2020efficient}:} A representative clustered FL algorithm that groups clients iteratively but lacks intra-cluster dynamic regularization.
\end{itemize}

For all FL methods, the backbone architecture for the local models is ResNet-18. The global communication rounds are set to $T=20$, and the number of local training epochs per round is $E=4$. 
For all FL scenarios, we investigate the influence of the number of clusters by setting $M \in \{1, 2, 3\}$. The configuration $M=1$ corresponds to a non-clustered baseline, equivalent to standard global aggregation. It should be noted that as IFCA is an inherently cluster-based algorithm designed to partition clients, it is only evaluated under $M=2$ and $M=3$ settings, whereas other baselines and the proposed Clustered-DRAPR are verified across all values of $M$ to demonstrate their adaptability.
For FedProx and our proposed method, the base proximal regularization hyperparameter is empirically set to $\mu_0 = 0.001$. All models are optimized using Stochastic Gradient Descent (SGD) with a uniform learning rate.

\begin{table*}[htbp]
\centering
\caption{Quantitative evaluation results (Accuracy, Precision, Recall, and F1-score \%) for multi-label classification under 5-class and 20-class settings. Results are presented as mean $\pm$ standard deviation.}
\label{tab:quant_results}
\resizebox{\textwidth}{!}{%
\renewcommand{\arraystretch}{1.2}
\begin{tabular}{ll cccc c cccc}
\toprule
\multirow{2}{*}{\textbf{Method}} & \multirow{2}{*}{\textbf{Number of Clusters ($M$)}} & \multicolumn{4}{c}{\textbf{5 Labels ($N_c=5$)}} & & \multicolumn{4}{c}{\textbf{20 Labels ($N_c=20$)}} \\
\cmidrule{3-6} \cmidrule{8-11}
 & & \textbf{Accuracy} & \textbf{Precision} & \textbf{Recall} & \textbf{F1} & & \textbf{Accuracy} & \textbf{Precision} & \textbf{Recall} & \textbf{F1} \\
\midrule
Local & / & $37.83 \pm 0.91$ & $64.79 \pm 1.24$ & $77.87 \pm 3.37$ & $70.70 \pm 2.18$ & & $41.44 \pm 4.95$ & $45.01 \pm 5.97$ & $60.79 \pm 4.62$ & $51.72 \pm 5.43$ \\
\midrule
\multirow{3}{*}{FedAvg} 
 & 1 (no clustering) & $61.89 \pm 1.03$ & $79.81 \pm 0.42$ & $68.23 \pm 1.75$ & $72.53 \pm 1.38$ & & $38.71 \pm 3.12$ & $66.06 \pm 1.86$ & $44.65 \pm 5.22$ & $53.28 \pm 3.94$ \\
 & 2 & $62.74 \pm 0.35$ & $79.90 \pm 0.17$ & $69.53 \pm 0.29$ & $73.58 \pm 0.05$ & & $39.24 \pm 6.52$ & $65.78 \pm 1.36$ & $45.82 \pm 4.72$ & $54.01 \pm 5.28$ \\
 & 3 & $62.80 \pm 0.26$ & $79.77 \pm 0.01$ & $69.93 \pm 0.28$ & $73.76 \pm 0.21$ & & $39.49 \pm 6.67$ & $64.63 \pm 5.93$ & $47.42 \pm 3.05$ & $54.70 \pm 4.87$ \\
\midrule
\multirow{3}{*}{FedProx} 
 & 1 (no clustering) & $62.22 \pm 0.25$ & $79.28 \pm 0.42$ & $69.75 \pm 0.93$ & $73.49 \pm 0.36$ & & $39.98 \pm 5.25$ & $65.90 \pm 3.54$ & $47.94 \pm 4.43$ & $55.50 \pm 4.27$ \\
 & 2 & $62.35 \pm 0.25$ & $78.76 \pm 0.01$ & $71.20 \pm 0.41$ & $74.31 \pm 0.23$ & & $40.53 \pm 1.92$ & $65.39 \pm 5.23$ & $49.26 \pm 3.46$ & $56.19 \pm 1.05$ \\
 & 3 & $62.51 \pm 0.61$ & $78.19 \pm 0.32$ & $71.56 \pm 0.33$ & $74.22 \pm 0.27$ & & $40.69 \pm 4.33$ & $65.22 \pm 5.25$ & $49.29 \pm 3.86$ & $56.15 \pm 3.47$ \\
\midrule
\multirow{3}{*}{SCAFFOLD} 
 & 1 (no clustering) & $63.43 \pm 1.28$ & $79.87 \pm 1.31$ & $70.48 \pm 0.70$ & $74.13 \pm 0.01$ & & $\mathbf{42.77 \pm 2.83}$ & $64.11 \pm 2.97$ & $45.75 \pm 4.98$ & $53.40 \pm 3.57$ \\
 & 2 & $64.03 \pm 1.66$ & $79.56 \pm 1.56$ & $71.62 \pm 0.53$ & $74.47 \pm 0.21$ & & $39.43 \pm 4.25$ & $65.57 \pm 1.59$ & $46.19 \pm 3.86$ & $54.20 \pm 1.26$ \\
 & 3 & $63.69 \pm 1.73$ & $79.44 \pm 1.26$ & $71.29 \pm 0.36$ & $74.62 \pm 0.59$ & & $38.46 \pm 8.84$ & $65.19 \pm 2.25$ & $45.64 \pm 1.17$ & $53.69 \pm 4.45$ \\
\midrule
\multirow{3}{*}{MOON} 
 & 1 (no clustering) & $62.51 \pm 0.31$ & $76.62 \pm 0.16$ & $\mathbf{75.26 \pm 0.28}$ & $75.42 \pm 0.06$ & & $40.25 \pm 3.02$ & $64.79 \pm 8.96$ & $47.56 \pm 3.44$ & $54.85 \pm 3.65$ \\
 & 2 & $62.77 \pm 0.12$ & $77.06 \pm 0.22$ & $74.70 \pm 0.67$ & $75.46 \pm 0.53$ & & $40.37 \pm 1.35$ & $65.51 \pm 1.73$ & $47.68 \pm 1.96$ & $55.19 \pm 1.57$ \\
 & 3 & $62.79 \pm 0.45$ & $77.02 \pm 0.08$ & $\underline{75.25 \pm 0.40}$ & $75.69 \pm 0.04$ & & $\underline{42.30 \pm 4.22}$ & $64.63 \pm 1.96$ & $50.36 \pm 0.74$ & $56.61 \pm 3.13$ \\
\midrule
\multirow{2}{*}{IFCA} 
 & 2 & $62.56 \pm 0.28$ & $79.39 \pm 0.61$ & $70.58 \pm 0.93$ & $74.34 \pm 0.79$ & & $40.70 \pm 5.03$ & $65.57 \pm 2.26$ & $49.00 \pm 5.94$ & $56.09 \pm 3.67$ \\
 & 3 & $62.77 \pm 0.00$ & $79.20 \pm 0.00$ & $70.94 \pm 0.01$ & $74.52 \pm 0.46$ & & $40.52 \pm 5.08$ & $65.47 \pm 4.75$ & $49.17 \pm 3.97$ & $56.16 \pm 3.88$ \\
\midrule
\multirow{3}{*}{\textbf{Ours}} 
 & 1 (no clustering) & $\underline{66.44 \pm 1.97}$ & $77.28 \pm 3.25$ & $74.58 \pm 2.18$ & $75.61 \pm 0.08$ & & $40.30 \pm 3.43$ & $65.75 \pm 5.54$ & $48.23 \pm 5.89$ & $55.64 \pm 3.68$ \\
 & 2 & $63.95 \pm 2.23$ & $\mathbf{80.68 \pm 0.88}$ & $72.23 \pm 1.51$ & $\underline{75.95 \pm 0.18}$ & & $40.74 \pm 3.17$ & $\underline{67.50 \pm 2.04}$ & $\underline{51.92 \pm 1.72}$ & $\underline{58.69 \pm 2.75}$ \\
 & 3 & $\mathbf{66.95 \pm 1.80}$ & $\underline{80.38 \pm 0.21}$ & $74.06 \pm 0.86$ & $\mathbf{76.85 \pm 0.91}$ & & $40.66 \pm 1.75$ & $\mathbf{68.11 \pm 2.57}$ & $\mathbf{52.13 \pm 2.62}$ & $\mathbf{59.06 \pm 2.51}$ \\
\bottomrule
\end{tabular}%
}
\end{table*}

\begin{table*}[htbp]
\centering
\caption{Ablation study results verifying the individual contributions of the macro-level dynamic clustering ($M$) and the micro-level DRAPR module.}
\label{tab:ablation_results}
\resizebox{\textwidth}{!}{%
\renewcommand{\arraystretch}{1.2}
\begin{tabular}{ll cccc c cccc}
\toprule
\multirow{2}{*}{\textbf{Method Configuration}} & \multirow{2}{*}{\textbf{Number of Clusters ($M$)}} & \multicolumn{4}{c}{\textbf{5 Labels ($N_c=5$)}} & & \multicolumn{4}{c}{\textbf{20 Labels ($N_c=20$)}} \\
\cmidrule{3-6} \cmidrule{8-11}
 & & \textbf{Accuracy} & \textbf{Precision} & \textbf{Recall} & \textbf{F1} & & \textbf{Accuracy} & \textbf{Precision} & \textbf{Recall} & \textbf{F1} \\
\midrule
\multirow{3}{*}{w/o DRAPR} 
 & 1 (no clustering) & $62.16 \pm 0.43$ & $76.34 \pm 0.42$ & $68.85 \pm 0.86$ & $72.40 \pm 0.36$ & & $39.79 \pm 5.33$ & $63.84 \pm 3.45$ & $46.64 \pm 4.82$ & $53.90 \pm 4.27$ \\
 & 2 & $62.62 \pm 0.25$ & $76.87 \pm 0.14$ & $72.03 \pm 0.41$ & $74.37 \pm 0.23$ & & $40.23 \pm 1.99$ & $64.39 \pm 5.18$ & $47.46 \pm 3.49$ & $54.64 \pm 1.05$ \\
 & 3 & $62.57 \pm 0.61$ & $76.72 \pm 0.28$ & $71.71 \pm 0.47$ & $74.13 \pm 0.27$ & & $40.59 \pm 4.43$ & $64.92 \pm 5.14$ & $47.78 \pm 3.72$ & $55.05 \pm 3.47$ \\
\midrule
\multirow{3}{*}{\textbf{w/ DRAPR (Ours)}} 
 & 1 (no clustering) & $66.44 \pm 1.97$ & $77.28 \pm 3.25$ & $\mathbf{74.58 \pm 2.18}$ & $75.61 \pm 0.08$ & & $40.30 \pm 3.43$ & $65.75 \pm 5.54$ & $48.23 \pm 5.89$ & $55.64 \pm 3.68$ \\
 & 2 & $63.95 \pm 2.23$ & $\mathbf{80.68 \pm 0.88}$ & $72.23 \pm 1.51$ & $75.95 \pm 0.08$ & & $\mathbf{40.74 \pm 3.17}$ & $67.50 \pm 2.04$ & $51.92 \pm 1.72$ & $58.69 \pm 2.75$ \\
 & 3 & $\mathbf{66.95 \pm 1.80}$ & $80.38 \pm 0.21$ & $74.06 \pm 0.86$ & $\mathbf{76.85 \pm 0.91}$ & & $40.66 \pm 1.75$ & $\mathbf{68.11 \pm 2.57}$ & $\mathbf{52.13 \pm 2.62}$ & $\mathbf{59.06 \pm 2.51}$ \\
\bottomrule
\end{tabular}%
}
\end{table*}

\subsection{Quantitative Results and Analysis}
\label{sec:quantitative_results}

Table \ref{tab:quant_results} presents a comprehensive quantitative comparison between the proposed Clustered-DRAPR framework and five state-of-the-art FL baselines across both the 5-label ($N_c=5$) and 20-label ($N_c=20$) structural damage classification tasks. Under the moderately heterogeneous 5-class setting, traditional methods like SCAFFOLD and IFCA show reasonable performance, but the proposed framework consistently outperforms all baselines across all evaluated metrics. The performance gap widens drastically under the extreme ``double heterogeneity" of the 20-class setting. As the label distribution becomes severely skewed and the physical divergence between clients intensifies, baseline methods suffer from severe client drift and catastrophic forgetting, evidenced by a sharp decline in their predictive capabilities. In contrast, Clustered-DRAPR maintains exceptional robustness and yields a clear margin of superiority, unequivocally demonstrating its capability to handle complex, real-world infrastructural data heterogeneity.

\begin{figure*}[htbp]
    \centering
    \includegraphics[width=\textwidth]{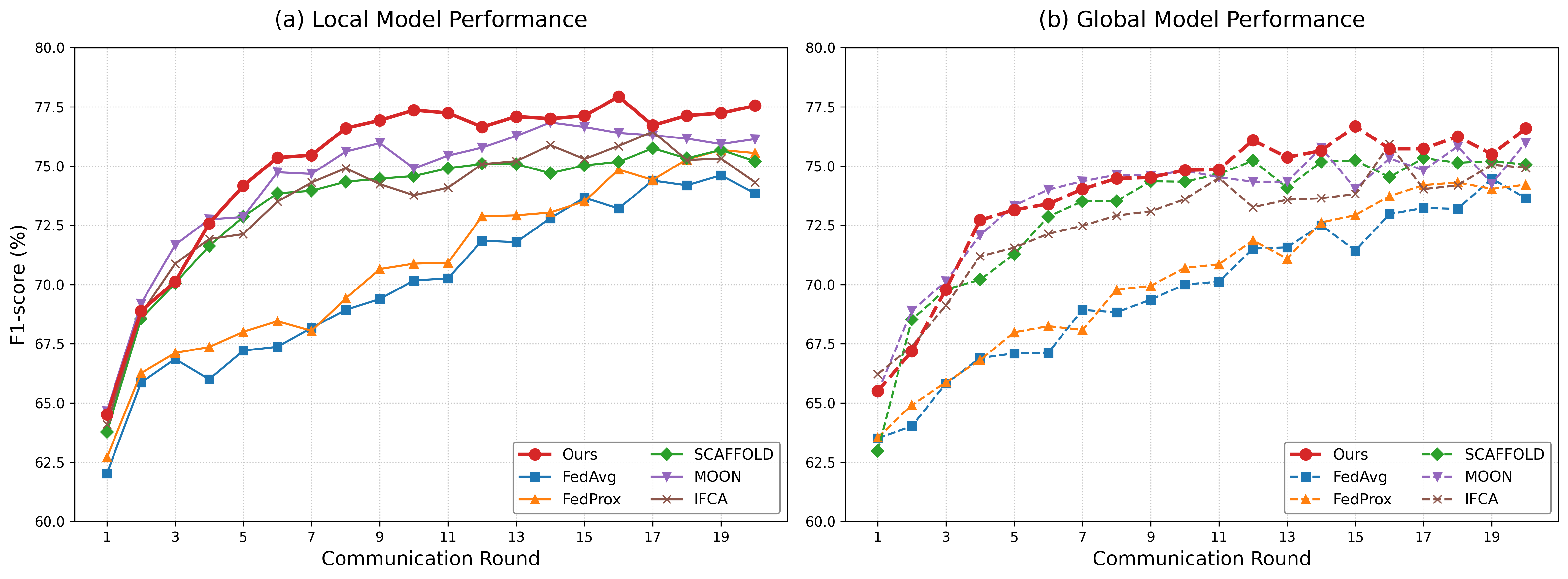}
    \caption{Learning curves illustrating the F1-score progression of various FL methods over 20 communication rounds under the 5-class setting. (a) Local model performance prior to server aggregation. (b) Global model performance after aggregation. The proposed Clustered-DRAPR framework (Ours) demonstrates faster convergence and superior stability in both local adaptation and global generalization compared to all baselines, effectively mitigating performance fluctuations caused by client drift.}
    \label{fig:f1_progression}
\end{figure*}

To further evaluate the training efficiency and algorithmic stability, Figure \ref{fig:f1_progression} traces the F1-score progression of the models over 20 communication rounds under the 5-class setting. The learning curves clearly illustrate that the proposed framework (Ours) achieves faster convergence and maintains a higher, more stable performance plateau in both local model adaptation (Figure \ref{fig:f1_progression}a) and global model generalization (Figure \ref{fig:f1_progression}b). By adaptively calibrating the proximal penalty based on real-time skewness, Clustered-DRAPR effectively mitigates the severe intra-round performance fluctuations that frequently destabilize standard methods like FedAvg and FedProx.

Finally, a dedicated ablation study was conducted to rigorously validate the individual contributions of the two core architectural modules: the macro-level dynamic clustering and the micro-level dynamic region-adaptive proximal regularization. As detailed in Table \ref{tab:ablation_results}, eliminating the macro-clustering mechanism (i.e., setting $M=1$) forces all structurally diverse clients to optimize toward a single, generalized global model. This non-clustered configuration results in a noticeable performance degradation compared to clustered arrangements ($M>1$), particularly when handling the complex 20-class scenario. This phenomenon confirms the fundamental necessity of macro-level clustering; micro-level regularization alone is insufficient to overcome inherent macroscopic physical divergence. Conversely, disabling the micro-level regularization (w/o DRAPR) under clustered settings degrades the intra-cluster optimization, leading to a consistent drop in overall metrics. This confirms that without the dynamic, $S_k^{(t)}$-driven penalty, the framework cannot adequately suppress local drift caused by microscopic statistical imbalance. Consequently, the ablation analysis conclusively demonstrates that neither component functions optimally in isolation; it is the synergistic, hierarchical integration of both mechanisms that establishes the framework's state-of-the-art robustness.

\begin{figure*}[htbp]
    \centering
    \includegraphics[width=\textwidth]{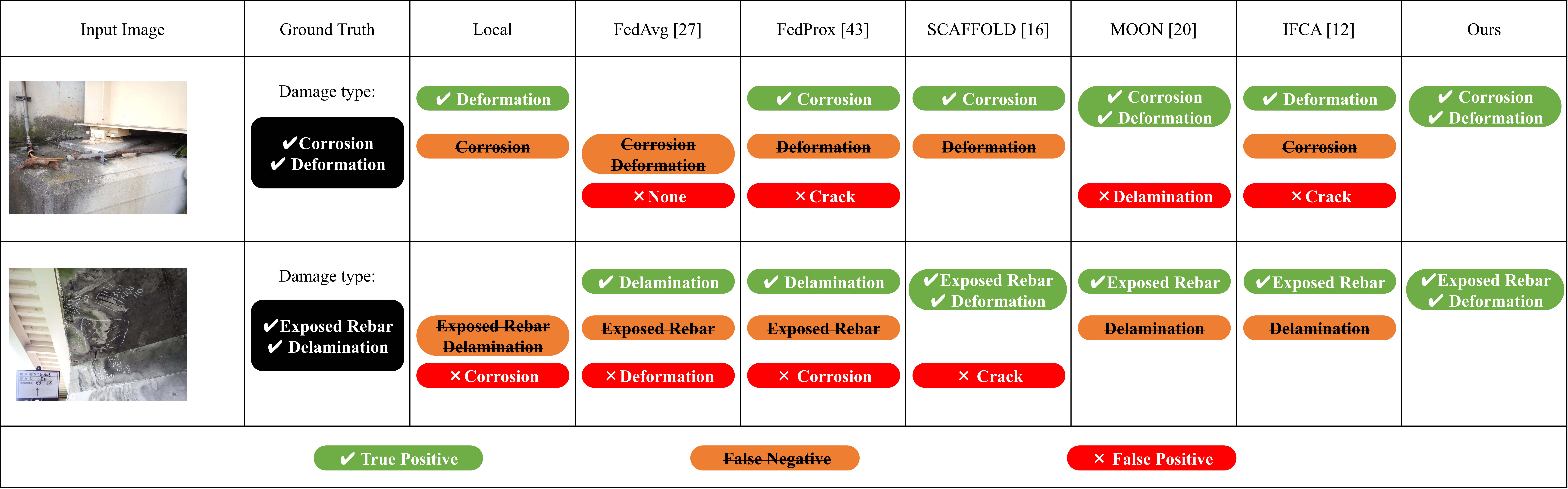} 
    \caption{Qualitative comparison of multi-label structural damage classification across various FL baselines. To enhance visual clarity, predictions are presented as functional tags: correctly predicted damages (true positives) are highlighted in \textcolor{mygreen}{green} with a checkmark, missed underlying damages (false negatives) are indicated in \textcolor{orange}{orange} with a strikethrough, and hallucinated predictions (false positives) are flagged in \textcolor{red}{red} with a cross. Compared to traditional methods that suffer from severe client drift---leading to both missing critical defects and generating false alarms---the proposed Clustered-DRAPR framework (Ours) precisely identifies the complete set of ground truth labels without any erroneous predictions.}
    \label{fig:qual_eval}
\end{figure*}

\subsection{Qualitative Evaluation and Visualization}
\label{sec:qualitative_evaluation}

To provide a comprehensive understanding of the framework's diagnostic capabilities and internal mechanisms, we present a dual-perspective qualitative evaluation focusing on both micro-level prediction accuracy and macro-level clustering interpretability.

First, Figure \ref{fig:qual_eval} illustrates the multi-label classification results on complex real-world structural images. To facilitate a clear visual comparison, predictions are encoded using functional tags: correctly identified damages (true positives) are marked in green, missed underlying defects (catastrophic forgetting/false negatives) are denoted by orange tags with a strikethrough, and hallucinated predictions (client drift/false positives) are highlighted in red. The visualizations explicitly reveal that traditional FL baselines frequently suffer from catastrophic forgetting (failing to detect co-occurring damages like "Deformation" or "Delamination") and overfit to local majority classes (hallucinating damages like "Crack"). In stark contrast, the proposed Clustered-DRAPR framework precisely identifies the complete set of ground truth labels without triggering any erroneous predictions, confirming that the $S_k^{(t)}$-driven regularization effectively preserves minority class knowledge and suppresses localized noise.

\begin{figure*}[htbp]
    \centering
    \begin{subfigure}[b]{0.48\textwidth}
        \centering
        \includegraphics[width=\textwidth]{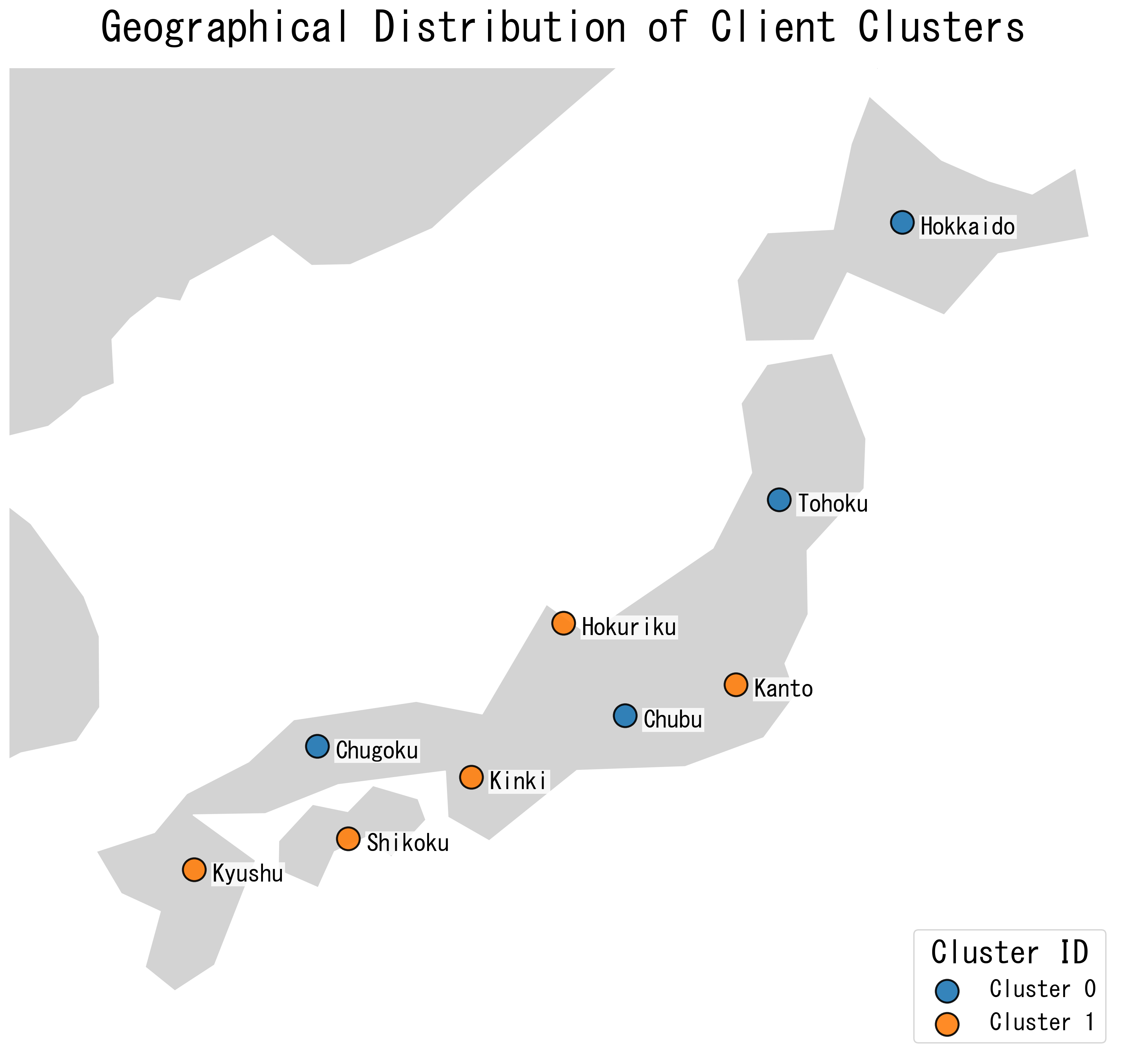}
        \caption{Dynamic clustering result with $M=2$}
        \label{fig:cluster_k2}
    \end{subfigure}
    \hfill
    \begin{subfigure}[b]{0.48\textwidth}
        \centering
        \includegraphics[width=\textwidth]{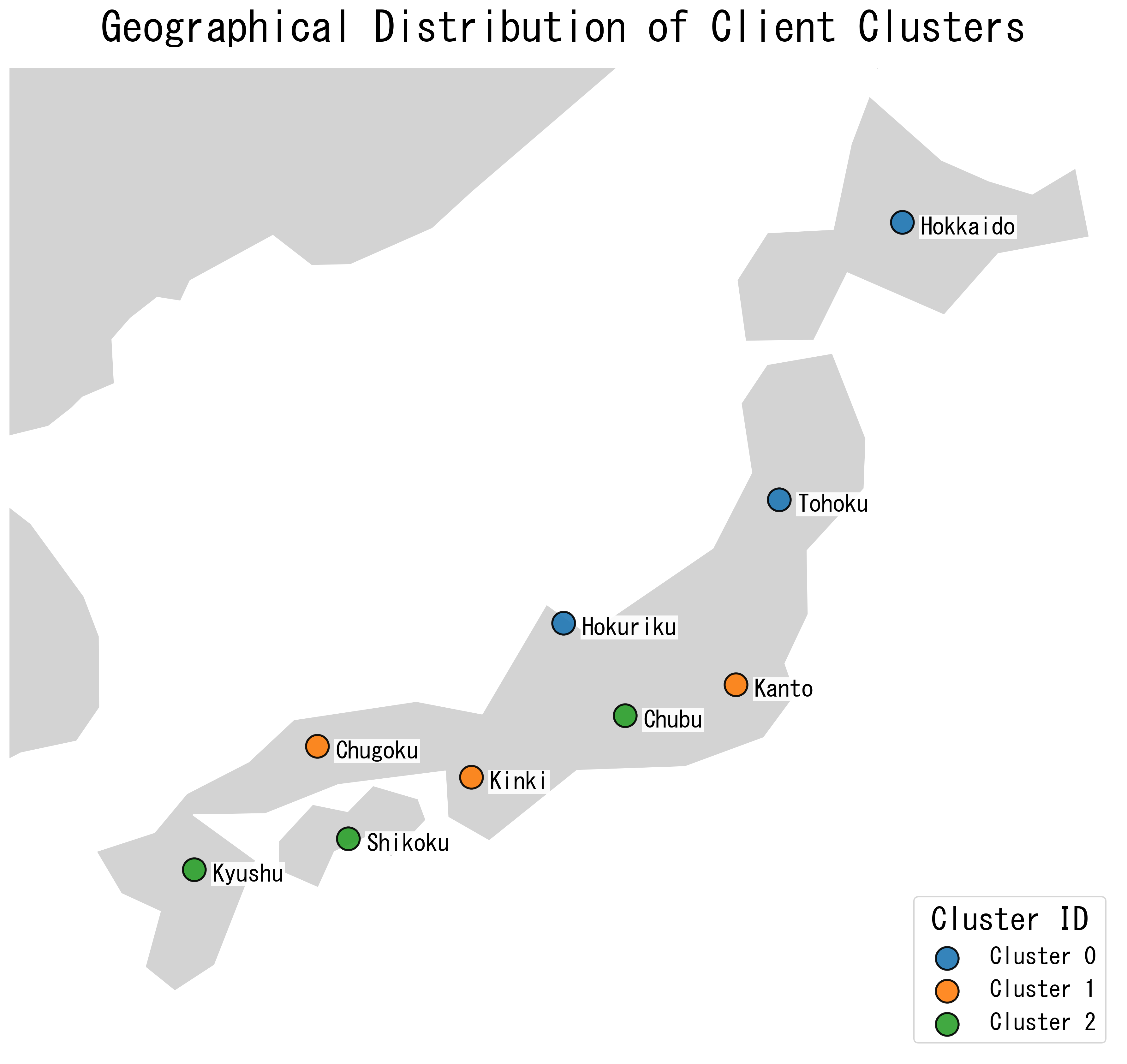}
        \caption{Dynamic clustering result with $M=3$}
        \label{fig:cluster_k3}
    \end{subfigure}
    \caption{Geographical visualization of the dynamically formed client clusters at communication round 20. The proposed framework aggregates clients based solely on the cosine similarity of their local model updates (gradients) without any prior geographical metadata. (a) Under the $M=2$ setting, the algorithm partitions the regional clients into two broad groups. (b) Under the $M=3$ setting, the clustering resolves into three distinct groups. This geographical alignment confirms that the gradient-matching strategy successfully captures underlying data distribution patterns.}
    \label{fig:cluster_maps}
\end{figure*}

Furthermore, the qualitative superiority and interpretability of the framework are corroborated by analyzing its macro-level clustering behavior. Figure \ref{fig:cluster_maps} visualizes the geographical distribution of the dynamically formed client clusters at communication round 20. Despite receiving no prior geographical or climatic metadata, the algorithm organically partitions the nine Japanese regional clients into distinct groups based solely on the cosine similarity of their local gradient updates. Under the $M=2$ setting (Figure \ref{fig:cluster_maps} (a)), the framework effectively separates the regions into two broad groups. More notably, under the optimal $M=3$ setting (Figure \ref{fig:cluster_maps} (b)), the clustering resolves into three distinct spatial groups: Cluster 0 (Hokkaido, Tohoku, and Hokuriku), Cluster 1 (Kanto, Kinki, and Chugoku), and Cluster 2 (Chubu, Shikoku, and Kyushu). 
This organic geographical alignment powerfully demonstrates that the gradient-matching strategy successfully captures the underlying data distribution patterns driven by diverse physical environments. Together, the precise image-level predictions and the highly interpretable spatial clustering visually confirm that the synergistic integration of macro-clustering and micro-regularization successfully neutralizes the double heterogeneity in large-scale infrastructure monitoring.

\section{Discussion}
\label{sec:discussion}

The proposed Clustered-DRAPR framework demonstrates a highly effective approach to overcoming the ``double heterogeneity" inherent in nationwide structural health monitoring (SHM). A profound finding of this research lies in the behavior of the macro-level dynamic clustering mechanism. Without utilizing any prior geographical or environmental metadata, the algorithm organically identified latent physical degradation patterns, successfully grouping clients into clusters that impeccably mirror real-world macro-climatic zones in Japan. This indicates that aggregating clients based on model update trajectories successfully encapsulate underlying infrastructural degradation patterns driven by physical environments.

Furthermore, the success of the framework highlights the critical necessity of micro-level interventions. The DRAPR module acts as a localized stabilizer. By dynamically computing a Non-IID Intensity Score based on label skewness and gradient divergence, DRAPR adaptively penalizes local client drift. This mechanism successfully neutralizes intra-cluster statistical imbalances and prevents the catastrophic forgetting of minority damage classes, which is a pervasive issue in standard FL deployments for infrastructure inspection.

While the current framework presents a robust solution for vision-based SHM, certain limitations provide avenues for future exploration. First, the present study assumes synchronous communication rounds. In real-world edge deployments—such as remote bridges with unstable network connectivity—an asynchronous FL aggregation scheme may be required to prevent straggler bottlenecks. Second, while this research focused exclusively on 2D inspection imagery, future work will aim to extend the Clustered-DRAPR framework into a multi-modal FL system. Integrating computer vision with continuous time-series sensor data (e.g., vibrations and strains) will be crucial to achieving a more comprehensive, digital-twin-driven prognostic health management system for civil infrastructure.

\section{Conclusion}
\label{sec:conclusion}
The decentralized nature of modern SHM, coupled with stringent data privacy regulations, has necessitated a paradigm shift from centralized data aggregation to distributed machine learning. However, deploying FL across nationwide infrastructure networks reveals a critical bottleneck: the ``double heterogeneity" characterized by macro-level physical divergence and micro-level statistical imbalance. To surmount this challenge, this paper proposed \textbf{Clustered-DRAPR}, a novel hierarchical FL framework specifically engineered for large-scale, privacy-preserving infrastructure inspection. 

By orchestrating a two-tier optimization strategy, the proposed framework successfully decouples the complex Non-IID problem through gradient-based dynamic clustering at the macro-level and DRAPR at the micro-level. Extensive experiments utilizing 77,890 authentic inspection records from the Japanese national xROAD database demonstrated the unequivocal superiority of the proposed framework. In multi-label classification tasks encompassing up to 20 distinct structural damage types, Clustered-DRAPR consistently outperformed state-of-the-art baselines (including FedProx, SCAFFOLD, MOON, and IFCA) in terms of accuracy, precision, recall, and F1-score. Qualitative evaluations further confirmed its exceptional capability to minimize both false negatives and false positives in complex, real-world inspection imagery. Ultimately, the ablation analysis verified that the synergistic integration of both the macro-clustering and micro-regularization modules is indispensable for achieving optimal performance under extreme heterogeneity, paving the way for reliable and scalable AI-driven infrastructure maintenance.

\appendix
\section{List of Structural Damage Labels}
\label{appendix:labels}

Table \ref{tab:appendix_labels} details the comprehensive list of 20 structural damage categories evaluated in the multi-label classification tasks. These categories are derived from the standardized bridge inspection protocols of the Japanese National Road Facility Inspection Database (xROAD).

\begin{table}[htbp]
\centering
\caption{Comprehensive list of 20 structural damage labels used in the experiments.}
\label{tab:appendix_labels}
\renewcommand{\arraystretch}{1.2}
\begin{tabular}{cl}
\toprule
\textbf{Label ID} & \textbf{Damage Category } \\
\midrule
01 & Corrosion \\
02 & Crack \\
03 & Exposed Rebar \\
04 & Delamination \\
05 & Deformation \\
06 & Efflorescence \\
07 & Puddling \\
08 & Pavement Abnormality \\
09 & Fissure \\
10 & Surface Irregularity \\
11 & Debris Accumulation \\
12 & Coating Deterioration \\
13 & Deck Slab Cracking \\
14 & Loose \\
15 & Breakage \\
16 & Damage to Reinforcement Materials \\
17 & Deterioration \\
18 & Bearing Dysfunction \\
19 & Anchorage Abnormality \\
20 & Settlement / Movement / Tilting \\
\bottomrule
\end{tabular}
\end{table}

\appendix
\section{Declaration of generative AI and AI-assisted technologies in the manuscript preparation process}
\label{appendix:declaration of generative AI}

During the preparation of this work the author(s) used Gemini in order to proofread and polish the English phrasing of the manuscript. After using this tool/service, the author(s) reviewed and edited the content as needed and take(s) full responsibility for the content of the published article.

\printcredits

\bibliographystyle{cas-model2-names}

\bibliography{cas-refs}


\bio{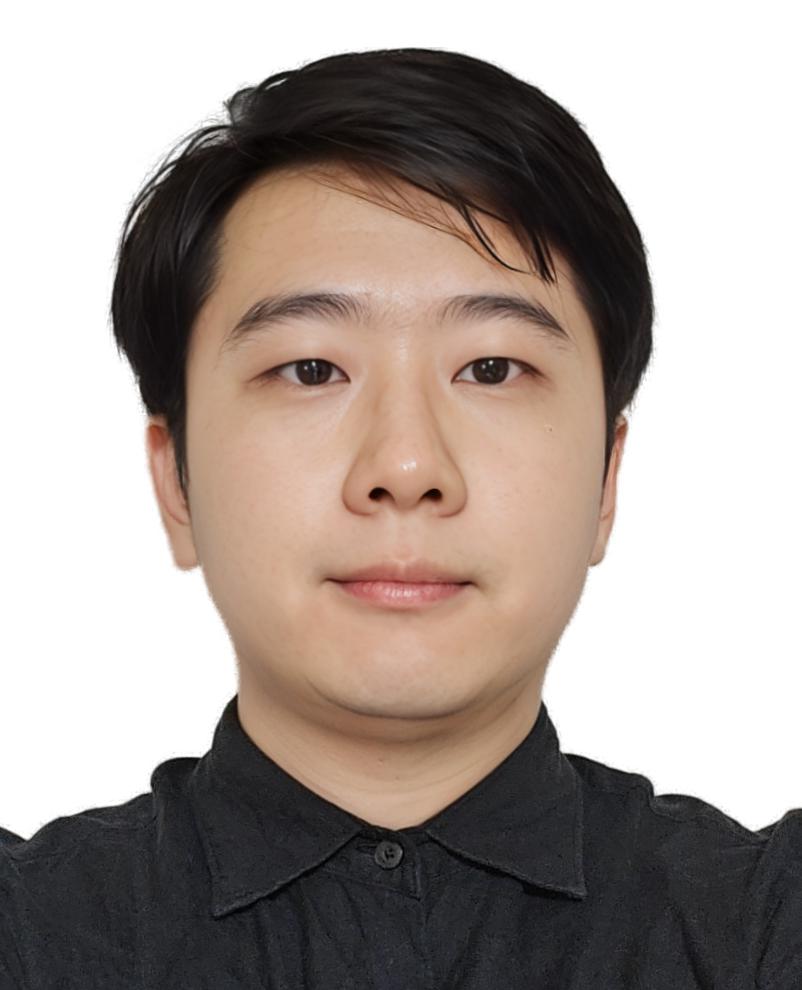}
(Graduate Student Member,IEEE) received the B.S. degree in Communication Engineering from Central South University, China, in 2020, and the M.S. degree in Information Science from Hokkaido University, Japan, in 2024. He is currently pursuing a Ph.D. degree with the Graduate School of Information Science and Technology at Hokkaido University. His research interests include image tetrieval and federated learning.
\endbio

\pagebreak

\bio{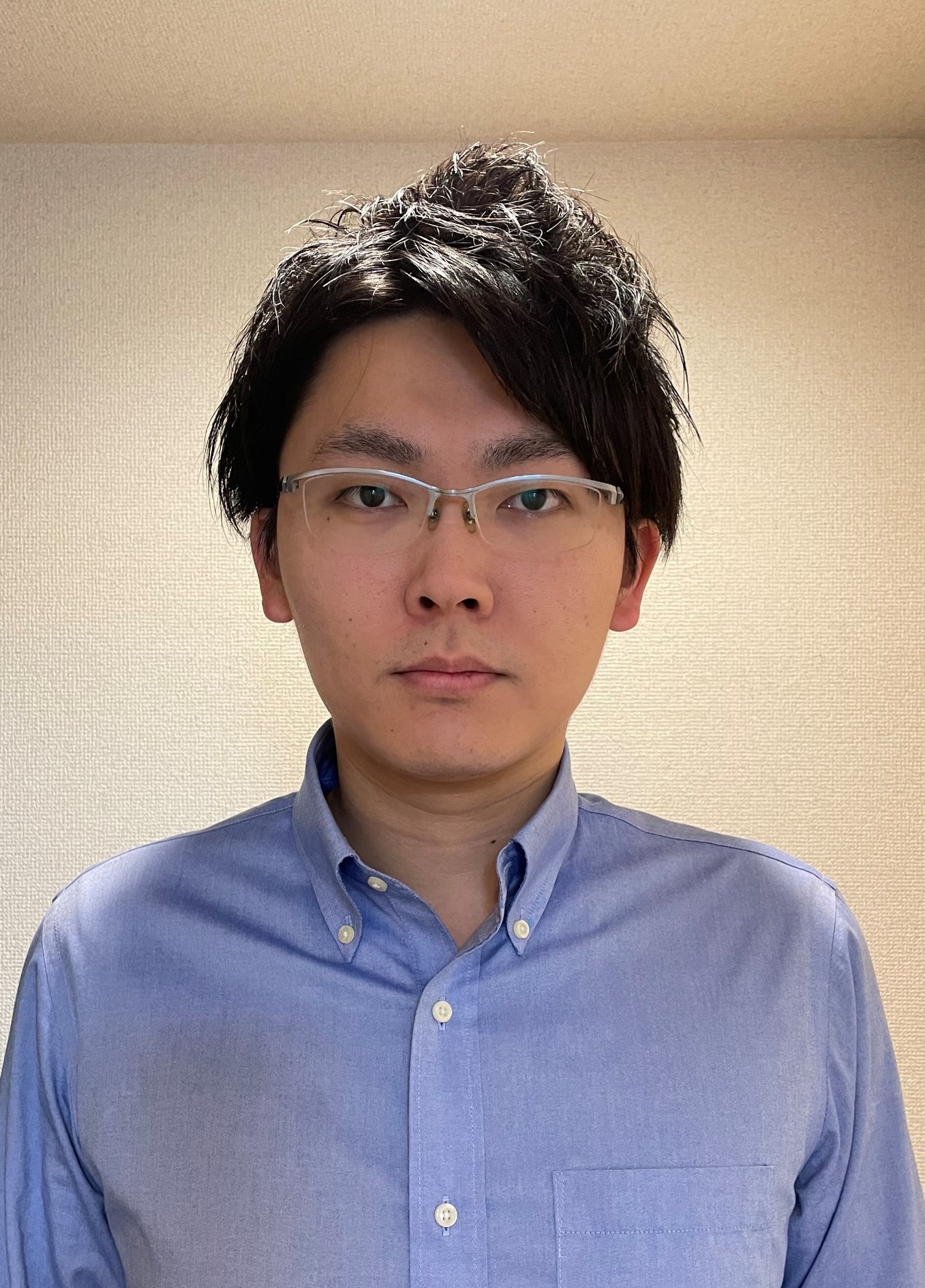}
(Member, IEEE) received the B.S., M.S., and Ph.D. degrees in electronics and information engineering from Hokkaido University, Japan, in 2015, 2017, and 2019, respectively. He is currently an Assistant Professor with the Faculty of Information Science and Technology, Hokkaido University. His research interests include multimodal signal processing, machine learning and its applications. He was a TPC Member of IEEE GCCE2019. He is a member of IEICE. He was the Organized Session Co-Chair of IEEE GCCE2020.
\endbio

\bio{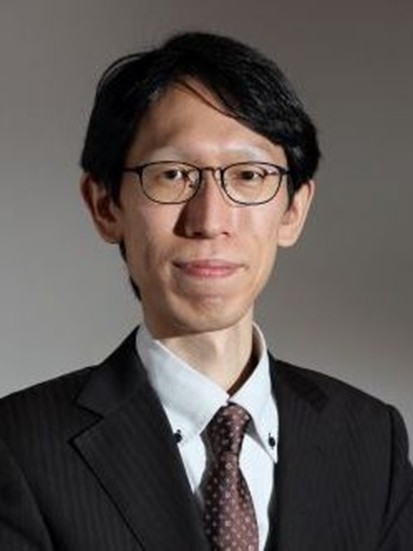}
(Senior Member, IEEE) received the B.S., M.S., and Ph.D. degrees in electronics and information engineering from Hokkaido University, Japan, in 2003, 2005, and 2007, respectively. He joined the Graduate School of Information Science and Technology, Hokkaido University, in 2008. He is currently a Professor with the Faculty of Information Science and Technology, Hokkaido University. His research interests include artificial intelligence, the Internet of Things, and big data analysis for multimedia signal processing and its applications. He has been the Special Session Chair of IEEE ISCE2009, the Doctoral Symposium Chair of ACM ICMR2018, the Organized Session Chair of IEEE GCCE2017-2019, the TPC Vice Chair of IEEE GCCE2018, and the Conference Chair of IEEE GCCE2019. He has also been an Associate Editor of ITE Transactions on Media Technology and Applications. He is a member of ACM, IEICE, and ITE.
\endbio

\bio{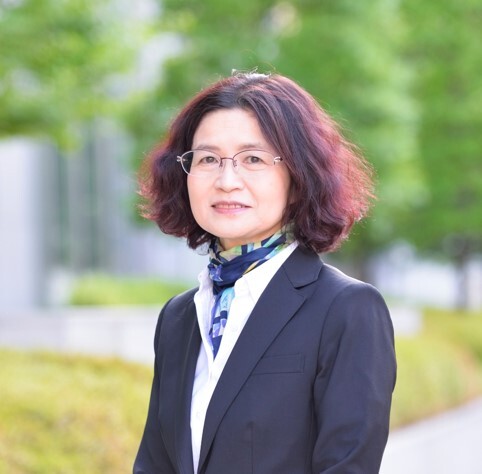}
(Senior Member, IEEE) received the B.S., M.S., and Ph.D. degrees in electronics from Hokkaido University, Japan, in 1986, 1988, and 1993, respectively. She joined the Graduate School of Information Science and Technology, Hokkaido University, as an Associate Professor, in 1994. She was a Visiting Associate Professor with Washington University, USA, from 1995 to 1996. She is currently a Professor with the Faculty of Information Science and Technology, Hokkaido University. Her research interests include image and video processing and its development into semantic analysis. She is a fellow of ITE and a member of IEICE and ASJ. She has been the Vice-President of the Institute of Image Information and Television Engineers, Japan (ITE); an Editor-in-Chief of ITE Transactions on Media Technology and Applications; the Director of International Coordination and Publicity at the Institute of Electronics, Information and Communication Engineers (IEICE).
\endbio

\end{document}